\documentstyle[epsbox,art10]{article}
\def\RIGHT{{\em RightNucleus}}
\def\LEFT{{\em LeftNucleus}}
\def\BOTH{{\em BothNucleus}}
\begin{document}

\twocolumn[%
\begin{center}
{\LARGE\bf ABSTRACT GENERATION \\
BASED ON \vspace*{1ex} \\
RHETORICAL STRUCTURE EXTRACTION} \vspace*{2ex}\\ \ \\
\large Kenji Ono, \ \ Kazuo Sumita, \ \ Seiji Miike\\
\vspace*{1ex}
\normalsize Research and Development Center\\
Toshiba Corporation\\
Komukai-Toshiba-cho 1, Saiwai-ku, Kawasaki, 210, Japan \\
ono@isl.rdc.toshiba.co.jp \vspace*{2ex} \\
\end{center}
\vspace*{6ex}
]

\section{ABSTRACT}\label{sec:abst}
We have developed an automatic abstract generation system
for Japanese expository writings based on
rhetorical structure extraction.
The system first extracts the rhetorical structure,
the compound of the rhetorical relations between sentences,
and then cuts out less important parts in the extracted structure
to generate an abstract of the desired length.\\
Evaluation of the generated abstract showed that
it contains at maximum 74\% of the most important sentences
of the original text.
The system is now utilized as a text browser for a prototypical interactive
document retrieval system.

\section{INTRODUCTION}\label{sec:intro}
Abstract generation is, like Machine Translation,
one of the ultimate goal of Natural Language Processing.
However, since conventional word--frequency--based abstract generation
systems(e.g. \cite{khn:aca})
are lacking in inter-sentential or discourse-structural analysis,
they are liable to generate incoherent abstracts.
On the other hand,
conventional knowledge or script--based abstract generation systems(e.g.
\cite{lehn:80}, \cite{fum:86}),
owe their success to the limitation of the domain,
and cannot be applied to document with varied subjects, such as
popular scientific magazine.
To realize a domain-independent abstract generation system, 
a computational theory for analyzing linguistic discourse structure
and its practical procedure
must be established.
%

Hobbs developed a theory in which he arranged three kinds of relationships
between sentences from the text coherency viewpoint \cite{hbb:cc}.

Grosz and Sidner proposed a theory which accounted for interactions between
three notions on discourse: linguistic structure, intention, and attention
\cite{grs:dsa}.

Litman and Allen described a model in which a discourse structure of
conversation was built by recognizing a participant's plans \cite{ltm:prms}.
These theories all depend on extra-linguistic knowledge, the accumulation of
which presents a problem in the realization of a practical analyzer.

Cohen proposed a framework for analyzing the structure of argumentative
discourse \cite{cohe:dsa}, yet did not provide a concrete identification
procedure for `evidence' relationships between sentences, where no linguistic
clues indicate the relationships. Also, since only relationships between
successive sentences were considered, the scope which the relationships cover
cannot be analyzed, even if explicit connectives are detected.

Mann and Thompson proposed a linguistic structure of text describing
relationships between sentences and their relative importance \cite{mnn:rst}.
However, no method for extracting the relationships from superficial linguistic
expressions was described in their paper.

We have developed a computational model of discourse for
Japanese expository writings, and
implemented a practical procedure for extracting
discourse structure\cite{sumi:dsa}.
In our model, discourse structure
is defined as the rhetorical structure, i.e.,
the compound of rhetorical relations between sentences in text.
Abstract generation is realized as a suitable application of the extracted
rhetorical structure.
In this paper we describe briefly our discourse model and
discuss the abstract generation system based on it.

\section{RHETORICAL STRUCTURE}\label{sec:d_str}

Rhetorical structure represents relations between various chunks of sentences
in the body of each section.
In this paper,
the rhetorical structure is represented by two layers:
intra--paragraph and inter-paragraph structures.
An intra--paragraph structure is a structure
whose representation units are sentences,
and an inter--paragraph structure is a structure
whose representation units are paragraphs.


In text, various rhetorical patterns are used to clarify the principle of
argument.
Among them, connective expressions, which state inter--sentence relationships,
are the most significant.
The typical grammatical categories of the connective expressions
are connectives and sentence predicates.
They can be divided into the thirty four categories
which are exemplified in Table~\ref{tbl:cr}.

\begin{table}[hbt]
\caption{Example of rhetorical relations}\label{tbl:cr}
\begin{center}\begin{tabular}{|c|l|} \hline\hline
Relation & Expressions \\ \hline\hline
serial (\verb+<SR>+) & {\it dakara} (thus) \\ \hline
summarization & {\it kekkyoku} (after all) \\
(\verb+<SM>+) & \\ \hline
negative (\verb+<NG>+) & {\it shikashi} (but) \\ \hline
example (\verb+<EG>+) & {\it tatoeba} (for example) \\ \hline
especial(\verb+<ES>+) & {\it tokuni} (particularly) \\ \hline
reason (\verb+<RS>+) & {\it nazenara} (because) \\ \hline
supplement (\verb+<SP>+) & {\it mochiron} (of course) \\ \hline
background (\verb+<BI>+) & {\it juurai} (hitherto) \\ \hline
parallel (\verb+<PA>+) & {\it mata} (and) \\ \hline
extension (\verb+<EX>+) & {\it kore wa} (this is) \\ \hline
rephrase (\verb+<RF>+) & {\it tsumari} (that is to say) \\ \hline
direction (\verb+<DI>+) & {\it kokode wa \ldots wo noberu} \\
                        & (here \ldots is described) \\ \hline\hline
\end{tabular}\end{center}
\end{table}

The rhetorical relation of a sentence,
which is the relationship to the preceding part of the text,
can be extracted in accordance with the connective expression in the sentence.
For a sentence without any explicit connective expressions,
extension relation is set to the sentence.
The relations exemplified in Table~\ref{tbl:cr} are used for representing
the rhetorical structure.

\begin{figure}[htb]
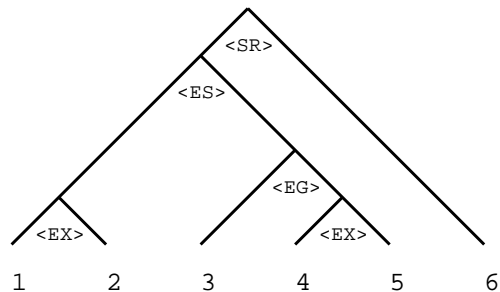

\begin{center}
\epsfile{file=dstruct.eps,height=40mm}
\end{center}
\caption{Rhetorical structure for the text in Fig.1 }
\label{fig:ds1}
\end{figure}

Fig.~\ref{fig:ex1} shows a paragraph from an article titled
``A Zero--Crossing Rate Which Estimates the Frequency of a Speech Signal,''
where underlined words indicate connective expressions.
Although the fourth and fifth sentences are clearly the exemplification of the
first three sentences, the sixth is not.
Also the sixth sentence is the concluding sentence for the first five.
Thus, the rhetorical structure for this text can be represented by a
binary--tree
as shown in Fig.~\ref{fig:ds1}.This structure is
also represented as follows:\vspace*{1ex}\\
\hspace*{1ex}{\small \verb+[[[1 <EX> 2] <ES> [3 <EG> [4 <EX> 5]]] <SR> 6]+ }
\vspace*{1ex}

\begin{figure}[hbt]
\begin{center}\begin{tabular}{|lp{7cm}|} \hline
1: & In the context of discrete--time signals, zero--crossing
is said to occur if successive samples have different algebraic signs. \\
2: & The rate at which zero crossings occur is a simple
measure of the frequency content of a signal. \\
3: & This is \underline{particularly} true of narrow band signals. \\
4: & \underline{For example}, a sinusoidal signal of frequency $F_0$,
sampled at a rate $F_s$, has $F_\phi / F_s$ samples per cycle of the sine wave.
\\
5: & Each cycle has two zero crossings so that the long--term
average rate of zero--crossings is $Z = 2 F_0 / F_s$. \\
6: & \underline{Thus}, the average zero--crossing rate gives a reasonable
way to estimate the frequency of a sine wave. \\
   & {\footnotesize (L.R.Rabiner and R.W.Schafer, {\em Digital Processing of
Speech Signals}, Prentice--Hall, 1978, p.127.)} \\ \hline
\end{tabular}\end{center}
\caption{Text example}
\label{fig:ex1}
\end{figure}

The rhetorical structure is represented by a binary tree
on the analogy of a syntactic tree of a natural language sentence.
Each sub tree of the rhetorical structure forms an argumentative constituent,
just as each sub--tree of the syntactic tree forms a grammatical
constituent.
Also, a sub--tree of the rhetorical structure is sub--categorized by a relation
of its parent node as well as a syntactic tree.

\section{RHETORICAL STRUCTURE EXTRACTION}\label{sec:d_strext}
The rhetorical structure represents logical relations between sentences
or blocks of sentences of each section of the document.
A rhetorical structure analysis determines logical relations between sentences
based on linguistic clues, such as connectives, anaphoric expressions,
and idiomatic expressions in the input text,
and then recognizes an argumentative chunk of sentences.

Rhetorical structure extraction
consists of six major sub--processes:

\begin{description}
\item[(1) Sentence analysis]
accomplishes morphological and syntactic analysis for each sentence.
\item[(2) Rhetorical relation extraction]
detects rhetorical relations and constructs the sequence of sentence
identifiers and relations.
\item[(3) Segmentation]
detects rhetorical expressions between distant sentences which
define rhetorical structure. They are added onto the sequence produced in
step 2, and form restrictions for generating structures in step 4.
For example, expressions like
``\ldots 3 reasons. First, \ldots \ \ Second,  \ldots \ \ Third, \ldots'', and
``\ldots \ \ Of course, \ldots \ \  \ldots But, \ldots '' are extracted and
the structural constraint is added onto the sequence
so as to form a chunk between the expressions.

\item[(4) Candidate generation]
generates all possible rhetorical structures described by binary trees which
do not violate segmentation restrictions.
\item[(5) Preference judgement]
selects the structure candidate with the lowest penalty score,
a value determined based on preference rules on every two neighboring
relations in the candidate.
This process selects the structure candidate with the lowest penalty score,
a value determined based on preference rules on every two neighboring
relations in the candidate.
A preference rule used in this process
represents a heuristic local preference on
consecutive rhetorical relations between sentences.
Consider the sequence
\verb+[P <EG> Q <SR> R]+, where \verb+P+, \verb+Q+, \verb+R+ are arbitrary
(blocks of) sentences.
The premise of \verb+R+ is obvously not only \verb+Q+ but both \verb+P+ and
\verb+Q+.
Since the discussion in \verb+P+ and \verb+Q+ is considered to close locally,
structure \verb+[[P <EG> Q] <SR> R]+ is preferable to
\verb+[P <EG> [Q <SR> R]]+.
Penalty scores are imposed on the structure candidates violating
the preference rules.
For example, for the text in Fig.~\ref{fig:ex1},
the structure candidates which contain the substructure \\
\verb+[3 <EG> [[4 <EX> 5] <SR> 6]]+ ,
which says
sentence six is the entailment of sentence four and five only,
are penalized.
The authors have investigated all pairs of rhetorical relations
and derived those preference rules.

\end{description}

The system analyzes inter--paragraph structures
after the analysis of intra--paragraph structures.
While the system uses the rhetorical relations of the first sentence
of each paragraph for this analysis,
it executes the same steps as it does for the intra--paragraph analysis.

\section{ABSTRACT GENERATION}
The system generates the abstract of each section of the document
by examining its rhetorical structure.
The process consists of the following 2 stages.
\begin{description}
\item[(1) Sentence evaluation]
\item[(2) Structure reduction]
\end{description}
In the {\it sentence evaluation} stage,
the system calculate the importance of each sentence in the original text
based on the relative importance of rhetorical relations.
They are categorized into three types as shown in Table~\ref{tbl:rimp}.
For the relations categorized into \RIGHT,
the right node is more important, from the point of view of abstract
generation, than the left node.
In the case of the \LEFT \ relations, the situation is vice versa.
And both nodes of the \BOTH \ relations are equivalent in their importance.
For example,
since the right node of the serial relation (e.g., {\it yotte} (thus))
is the conclusion of the left node,
the relation is categorized into \RIGHT,
and the right node is more important than the left node.

The Actual sentence evaluation is carried out in a demerit marking way.
In order to determine important text segments,
the system imposes penalties on both nodes for each rhetorical relation
according to its relative importance.
The system imposes a penalty on the left node for the \RIGHT \ relation,
and also on the right node for the \LEFT \ relation.
It adds penalties from the root node to the terminal nodes in turn,
to calculate the penalties of all nodes.

Then, in the {\it structure reduction} stage,
the system recursively cuts out the nodes, from the terminal nodes,
which are imposed the highest penalty.
The list of terminal nodes of the final structure becomes
an abstract for the original document.
Suppose that the abstract is longer than the expected length.
In that case the system cuts out terminal nodes from the last sentences,
which are given the same penalty score.

If the text is written loosely,
the rhetorical structure generally contains many \BOTH \ relations
(e.g., parallel({\it mata}(and, also)), and
the system cannot gradate the penalties and cannot reduce sentences smoothly.

After sentences of each paragraph are reduced,
inter-paragraph structure reduction is carried out in the same way
based on the relative importance judgement
on the inter-paragraph rhetorical structure.

\begin{figure}[htb]
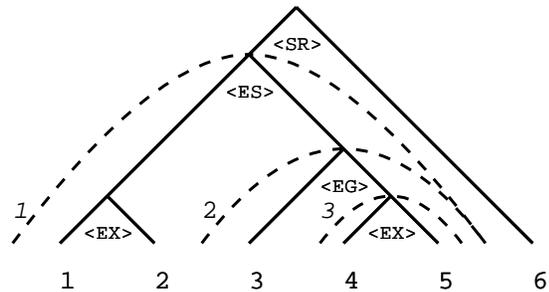

\begin{center}
\epsfile{file=relimp.eps,height=40mm}
\end{center}
\caption{Penalties on relative importance for the rhetorical structure in
Fig.2}\label{fig:pri}
\end{figure}

If the penalty calculation mentioned above is accomplished
for the rhetorical structure shown in Fig.~\ref{fig:ds1},
each penalty score is calculated as shown in Fig.~\ref{fig:pri}.
In Fig.~\ref{fig:pri} italic numbers are the penalties the system
imposed on each node of the structure,
and broken lines are the boundary between the nodes imposed different penalty
scores.
The figure shows
that sentence four and five have penalty score three,
that sentence three has two ,
that sentence one and two have one,
and that sentence six has no penalty score.
In this case,
the system selects sentence one, two, three and six
for the longest abstract, and
and also could select sentence one, two and six
as a shorter abstract,
and also could select sentence six
as a still more shorter abstract.


After the sentences to be included in the abstract are determined,
the system alternately arranges the sentences and the connectives
from which the relations were extracted, and
realizes the text of the abstract.

The important feature of the generated abstracts is that
since they are composed of
the rhetoricaly consistent units
which consist of several sentences and
form a rhetorical substructure,
the abstract does not contain fragmentary sentences
which cannot be understood alone.
For example, in the abstract generation mentioned above,
sentence two does not appear solely in the abstract, but
appears always with sentence one.
If sentence two appeared alone in the abstract without sentence one,
it would be difficult to understand the text.


\begin{table}[bt]
\caption{Relative importance of rhetorical relations}\label{tbl:rimp}
\begin{center}\begin{tabular}{|c|l|c|}\hline\hline
Relation Type & \multicolumn{1}{c|}{Relation} & Import. Node \\ \hline\hline
       & serial, & \\
\RIGHT & summariza- & right node \\
       & tion,      & \\
       & negative, \ldots & \\ \hline
       & example, & \\
\LEFT  & reason, & left node \\
       & especial, & \\
       & supplement, & \\
       & \ldots & \\ \hline
       & parallel, & \\
\BOTH  & extension, & both nodes \\
       & rephrase, \ldots & \\ \hline\hline
\end{tabular}\end{center}
\end{table}

\section{EVALUATION}\label{sec:eval}

The generated abstracts were evaluated
from the point of view of key sentence coverage.
30 editorial articles of "Asahi Shinbun", a Japanese newspaper, and
42 technical papers of "Toshiba Review", a journal of Toshiba Corp.
which publishes short expository papers of three or four pages,
were selected and
three subjects judged the key sentences and the most important key sentence
of each text.
As for the editorial articles,
The average correspondence rates of the key sentence and
the most important key sentence among the subjects were
60\% and 60\% respectively.
As for the technical papers,
they were 60\% and 80 \% respectively.

Then the abstracts were generated and
were compared with the selected key sentences.
The result is shown in Table~\ref{tbl:eval}.
As for the technical papers,
the average length ratio( abstract/original ) was 24 \%, and
the coverage of the key sentence and the most important key sentence were
51\% and 74\% respectively.
Whereas,
as for the editorials,
the average length ratio( abstract/original ) was 30 \%, and
the coverage of the key sentence and the most important key sentence were
41\% and 60\% respectively.

The reason why the compression rate and the key sentence coverage of
the technical papers were higher than
that of the editorials is  considered as follows.
The technical papers contains so many rhetorical expressions in general
as to be expository.
That is, they provide many linguistic clues
and the system can extract the rhetorical structure exactly.
Accordingly, the structure can be reduced further and
the length of the abstract gets shorter, without omitting key sentences.
On the other hand, in the editorials most of the relations between sentences
are supposed to be understood semantically, and are not expressed rhetorically.
Therefore, they lack linguistic clues
and the system cannot extract the rhetorical structure exactly.

\begin{table}[hbt]
\caption{ Key sentence coverage of the abstracts}
\label{tbl:eval}
\newcommand{\lw}[1]{\smash{\lower2.ex\hbox{#1}}}
\newcommand{\lv}[1]{\smash{\lower1.ex\hbox{#1}}}
\newcommand{\lx}[1]{\smash{\lower3.ex\hbox{#1}}}
\begin{center}
{\small
\begin{tabular}{|c|c|c|c|c|}
\hline
\lx{Material} & \lx{total}  & \lx{length} &
								\multicolumn{2}{c|}{ \lv{cover ratio} }\\
              & \lx{num.} & \lx{ratio} & \multicolumn{2}{c|}{ }\\
\cline{4-5}
 & & & key & most \\
 & & & sentence & important \\
 & & &          & sentence \\
\hline \hline

editorial & 30 & 0.3 & 0.41 & 0.60 \\
{\footnotesize \it (Asahi Shinbun) } & & & & \\
tech. journal & 42 & 0.24 & 0.51 & 0.74 \\
{\footnotesize \it (Toshiba Review) } & & & & \\
\hline
\end{tabular}
}
\end{center}
\end{table}

\section{CONCLUSION}\label{sec:conc}

We have developed an automatic abstract generation system
for Japanese expository writings based on
rhetorical structure extraction.

The rhetorical structure provides a natural order of importance among sentences
in the text,
and can be used to determine which sentence should be extracted in the
abstract,
according to the desired length of the abstract.
The rhetorical structure also provides the rhetorical relation between the
extracted sentences,
and can be used to generate appropriate connectives between them.

Abstract generation based on rhetorical structure extraction has four merits.
First, unlike conventional word--frequency--based abstract generation
systems(e.g. \cite{khn:aca}),
the generated abstract is consistent with the original text
in that the connectives between sentences in the abstract reflect their
relation in the original text.
Second, once the rhetorical structure is obtained,
various lengths of generated abstracts can be generated easily.
This can be done
by simply repeating the reduction process
until one gets the desired length of abstract.
Third, unlike conventional knowledge or script--based abstract generation
systems(e.g. \cite{lehn:80}, \cite{fum:86}),
the rhetorical structure extraction does not need prepared knowledge or scripts
related to the original text ,
and can be used for texts of any domain , so long as they contain enough
rhetorical expressions to be expository writings.
Fourth, the generated abstract is composed of
rhetoricaly consistent units
which consist of several sentences and
form a rhetorical substructure.
so the abstract does not contain fragmentary sentences
which cannot be understood alone.

The limitations of the system are mainly due to
errors in the rhetorical structure analysis
and the sentence-selection-type abstract generation.
the evaluation of the accuracy of the rhetorical structure analysis
carried out previously( \cite{sumi:dsa} )
showed 74\%.
Also, to make the length of the abstract shorter,
It is necessary to utilize an inner-sentence analysis and
to realize a phrase-selection-type abstract generation based on it.
The anaphora-resolution and the topic-supplementation must also be
realized in the analysis.

The system is now utilized as a text browser for a prototypical interactive
document retrieval system.


\end{document}